\documentclass[letterpaper, 10 pt, conference]{ieeeconf} 
\IEEEoverridecommandlockouts                              
\usepackage{graphics} % for pdf, bitmapped graphics files
\usepackage{amsmath} % assumes amsmath package installed
\usepackage{amssymb}  % assumes amsmath package installed
\usepackage{cite}
\usepackage{amsfonts}
\usepackage{bm}
\usepackage{breqn}
\renewcommand{\vec}{\mathbf}

\usepackage{hyperref}
\usepackage[percent]{overpic}
\usepackage{xcolor}
\newcommand{\insertYoutubeLink}{\url{https://youtu.be/roE1vxpEWfw}}

\usepackage{multicol}
\usepackage{subcaption}
\usepackage{graphicx}

\title{\LARGE \bf
Robust High-Speed Running for Quadruped Robots via \\ Deep Reinforcement Learning
}

\author{Guillaume Bellegarda, Yiyu Chen, Zhuochen Liu, and Quan Nguyen% 
\thanks{This work is supported by USC Viterbi School of Engineering.}
\thanks{The authors are with the Dynamic Robotics and Control Laboratory, University of Southern California (USC).
        {\tt\small bellegar,yiyuc, liuzhuoc,quann@usc.edu}}%
}

\begin{document}
\bstctlcite{MyBSTcontrol}
\maketitle
\thispagestyle{empty}
\pagestyle{empty}

%%%%%%%%%%%%%%%%%%%%%%%%%%%%%%%%%%%%%%%%%%%%%%%%%%%%%%%%%%%%%%%%%%%%%%%%%%%%%%%%
% Abstract
%%%%%%%%%%%%%%%%%%%%%%%%%%%%%%%%%%%%%%%%%%%%%%%%%%%%%%%%%%%%%%%%%%%%%%%%%%%%%%%%
\begin{abstract} 

Deep reinforcement learning has emerged as a popular and powerful way to develop locomotion controllers for quadruped robots. Common approaches have largely focused on learning actions directly in joint space, or learning to modify and offset foot positions produced by trajectory generators. Both approaches typically require careful reward shaping and training for millions of time steps, and with trajectory generators introduce human bias into the resulting control policies. In this paper, we present a learning framework that leads to the natural emergence of fast and robust bounding policies for quadruped robots. The agent both selects and controls actions directly in task space to track desired velocity commands subject to environmental noise including model uncertainty and rough terrain. We observe that this framework improves sample efficiency, necessitates little reward shaping, leads to the emergence of natural gaits such as galloping and bounding, and eases the sim-to-real transfer at running speeds. Policies can be learned in only a few million time steps, even for challenging tasks of running over rough terrain with loads of over 100\% of the nominal quadruped mass. Training occurs in PyBullet, and we perform a sim-to-sim transfer to Gazebo and sim-to-real transfer to the Unitree A1 hardware. For sim-to-sim, our results show the quadruped is able to run at over 4 m/s without a load, and 3.5 m/s with a 10 kg load, which is over 83\% of the nominal quadruped mass. For sim-to-real, the Unitree A1 is able to bound at 2 m/s with a 5 kg load, representing 42\% of the nominal quadruped mass. 
\end{abstract}

%%%%%%%%%%%%%%%%%%%%%%%%%%%%%%%%%%%%%%%%%%%%%%%%%%%%%%%%%%%%%%%%%%%%%%%%%%%%%%%%
% Introduction
%%%%%%%%%%%%%%%%%%%%%%%%%%%%%%%%%%%%%%%%%%%%%%%%%%%%%%%%%%%%%%%%%%%%%%%%%%%%%%%%
\section{Introduction}
\label{sec:introduction}

Traditional control methods have made significant advances toward real world robot locomotion \cite{dicarlo2018mpc,bellicoso2018dynamic,kim2019highly,sombolestan2021adaptive}. Such methods typically rely on solving online optimization problems (MPC) using simplified dynamics models, which require significant domain knowledge, and may not generalize to new environments not explicitly considered during development (i.e. uneven slippery terrain). There is also considerable bias in the resulting motions due to the (for example) pre-determined foot swing trajectories, and use of Raibert heuristics \cite{raibert1986legged}, which are based on a linear inverted pendulum model, for foot placement planning. Such biases raise questions on optimality, with respect to speed and energy efficiency. 

By contrast, to make control policies more robust to external disturbances and new unseen environments, deep learning has emerged as an attractive and generalizable formulation. Deep reinforcement learning in particular has recently shown impressive results in learning control policies for legged systems such as bipeds~\cite{xie2020cassie, siekmann2021sim, siekmann2021blind} and quadrupeds~\cite{hwangbo2019anymal, lee2020anymal,tan2018minitaur,peng2020laikagoimitation,iscen2018policies,rahme2020dynamics, ji2022concurrent}. Typically such methods train from scratch (i.e. use little or no prior information about the system) and rely on extensive simulation with randomized environment parameters before transferring to hardware. 

\begin{figure}[!tpb]
      \centering
      \includegraphics[width=0.493\linewidth,trim={0 0 0 2cm},clip]{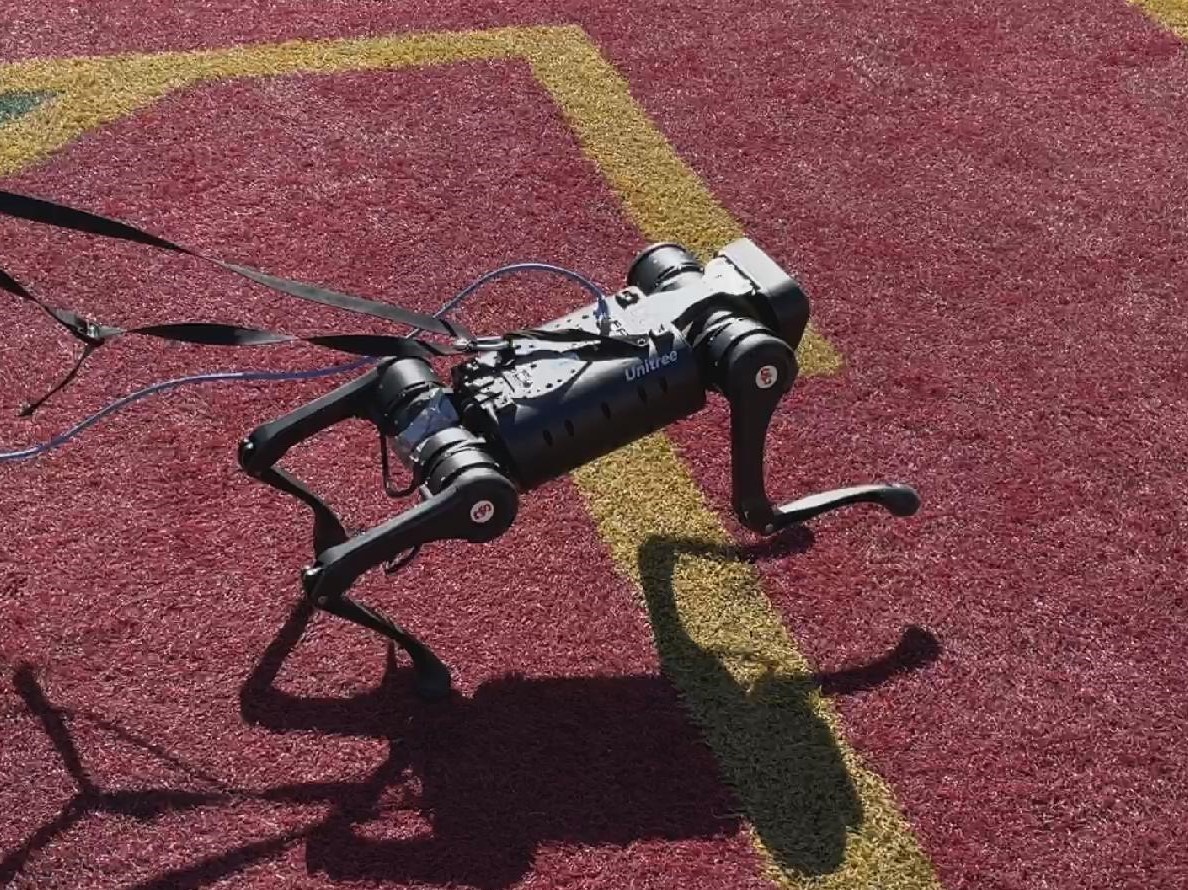}
      \includegraphics[width=0.493\linewidth,trim={0 0 0 2cm},clip]{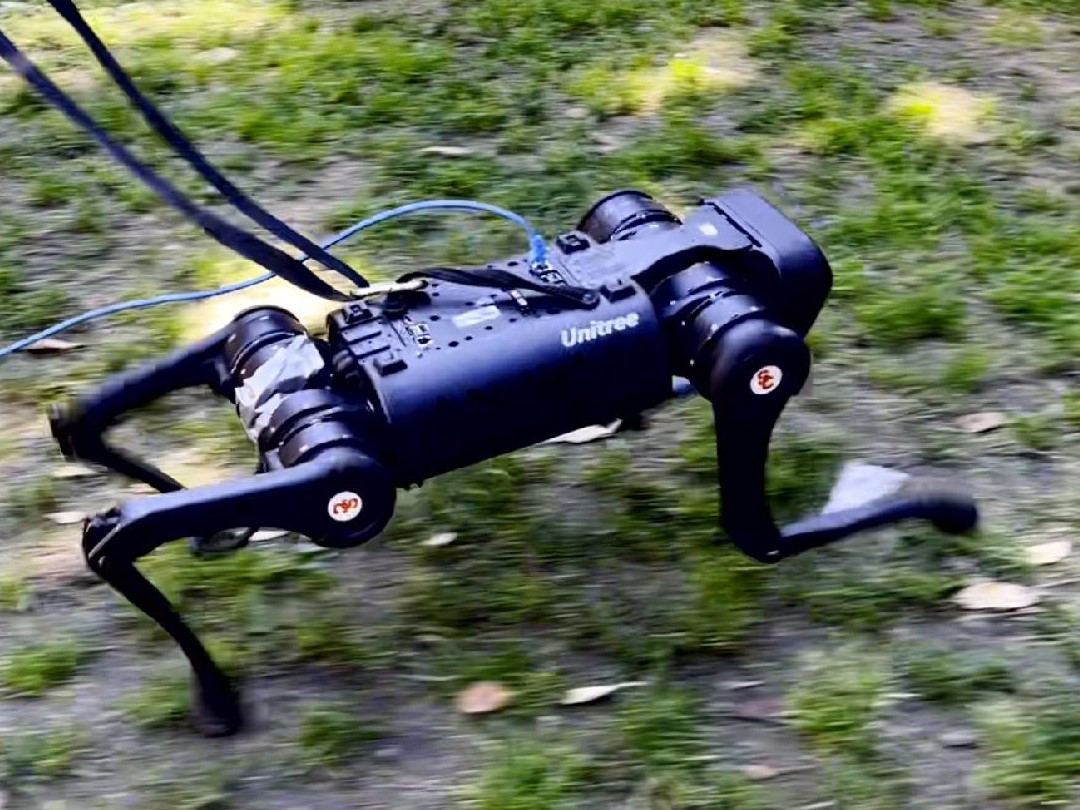} \\
      \vspace{0.008\linewidth}
      \includegraphics[width=0.493\linewidth]{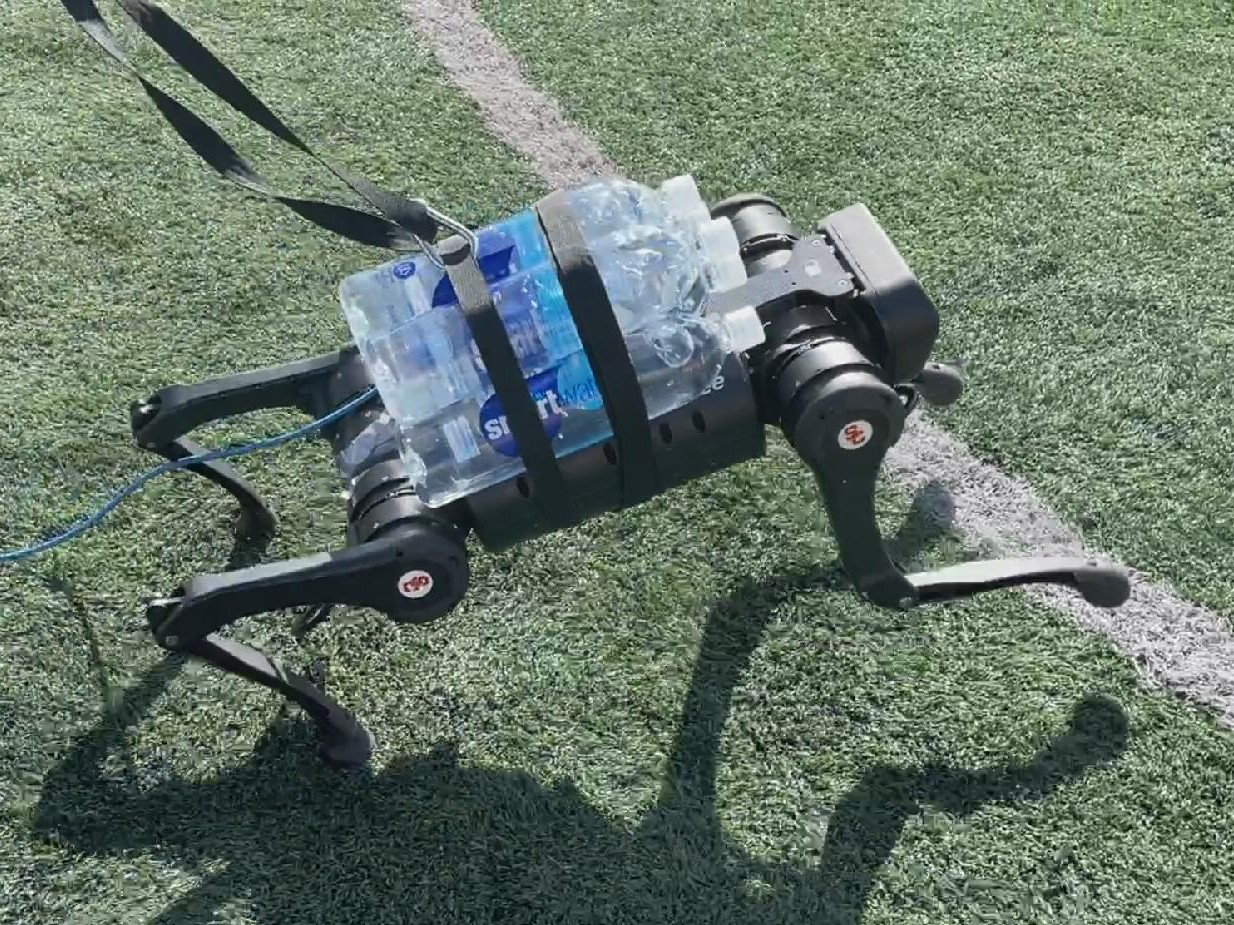}
    \reflectbox{\includegraphics[width=0.493\linewidth]{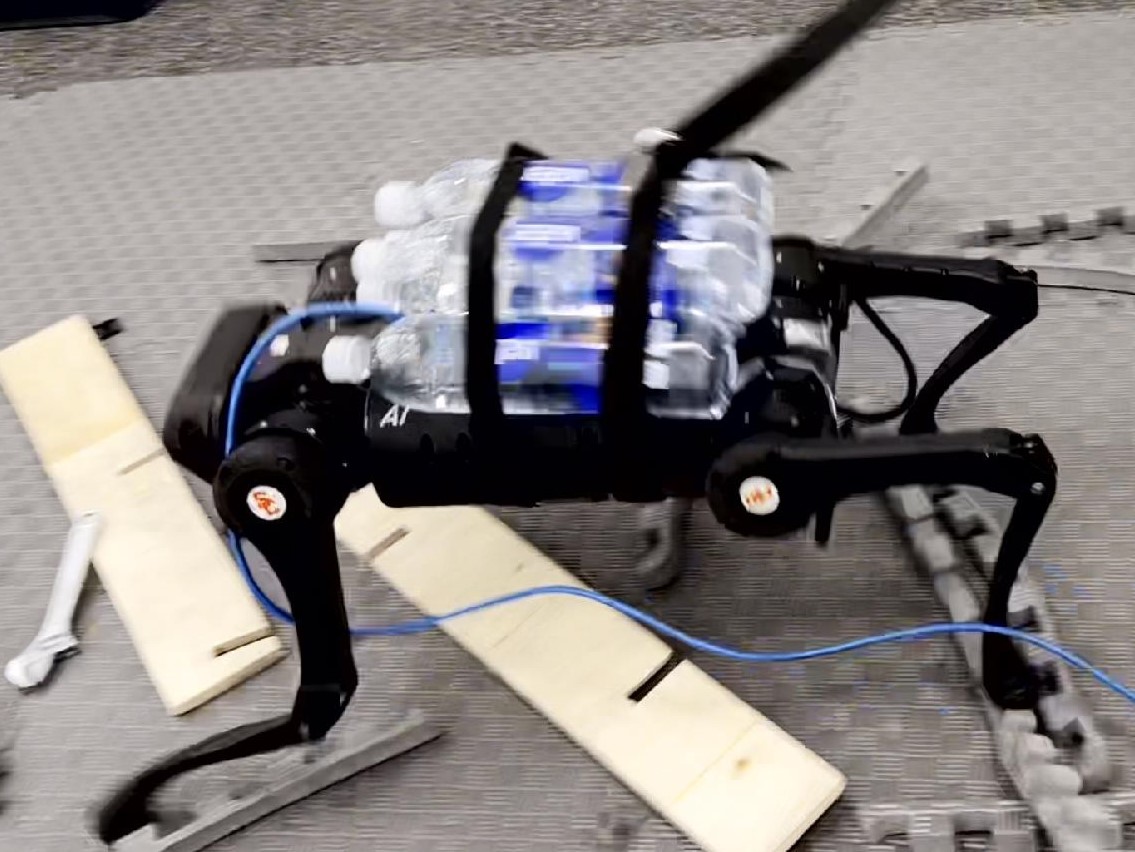}}
      \caption{Learning robust running on the Unitree A1 robot. 
    \textbf{Top:} Running at 2 $m/s$ at left, and over rough terrain at right. 
     \textbf{Bottom:} Running at 2 $m/s$ with a 5kg load (42\% of the robot's mass) at left, and over rough terrain at right. Experiment video link: \url{https://youtu.be/3gdRFzNQd68}.
      }
      \vspace{-1em}
      \label{fig:intro}
\end{figure}

Similarly to model-based control, recent works have shown successful results in learning ``blind'' control policies, with only proprioceptive sensing being mapped to joint commands. For quadrupedal robots, such approaches have resulted in successful locomotion controllers on both flat~\cite{tan2018minitaur,hwangbo2019anymal} and rough terrain~\cite{kumar2021rma,lee2020anymal}. Other works have shown the possibility of directly imitating animal motions~\cite{peng2020laikagoimitation}, and the emergence of different gaits through minimizing energy consumption with deep reinforcement learning~\cite{fu2022energy}. More recent works have integrated vision with proprioceptive sensing for tasks such as obstacle avoidance~\cite{yang2022learning} as well as dynamic crossing of rough terrain through the use of sampling from height maps~\cite{miki2022learning,rudin2022anymalisaac}. Additional works have shown gap crossing~\cite{yu2022visual}, also with full flight phases learned from pixels and leveraging MPC~\cite{margolis2022pixels}. 

Directly mapping from proprioceptive sensing to joint space commands typically involves careful reward shaping with many weighted terms, and on the order of millions or even billions of simulated training timesteps. To better structure the learning problem and avoid undesired local optima, recent works are exploring different action spaces, and in particular modulating outputs of high level trajectory generators~\cite{iscen2018policies}. This idea has lead to successful sim-to-real transfers for a variety of platforms including quadrupeds~\cite{lee2020anymal,miki2022learning} and bipeds~\cite{siekmann2021blind}. Additionally, incorporating phase encodings facilitates learning different gaits~\cite{shao2022gait}, and can also be used together with MPC~\cite{yang2022fast,xie2021glide}. 

With trajectory generators and MPC, there are questions about optimality with respect to both speed and efficiency, as the resulting policies are biased by design choices.  Other studies have also shown that action space choice has large effects on both training time and policy quality, for general robotic systems ranging from manipulators~\cite{martinIROS19VICES,stulp2012impedance,buchli2010impedanceRL,luo2019impedance} to legged systems~\cite{peng2017actionspace,bellegardaIROS19TaskSpaceRL,bellegarda2020robust, duan2021learning}. These works show that choosing the right action space is critical to learning an optimal policy, and in particular suggest that selecting actions in Cartesian space can outperform learning in joint space, depending on the task. 

The sim-to-real transfer is another important aspect of this problem. There are several methods to improve the robustness of policies trained in simulation, such as domain randomization \cite{tobin2017domain, exarchos2021policy} and dynamics randomization~\cite{peng2018sim, mordatch2015ensemble, xie2021dynamics}. In addition to sufficient simulation environmental noise or accurate dynamics modeling, policy representation in terms of neural network architecture also plays an important role for the sim-to-real transfer. Previous works\cite{siekmann2021blind, peng2018sim, siekmann2020learning} have shown that a memory-enabled network (i.e. LSTM) has higher proficiency to deal with partially observable environments. 

\textit{Problem Statement and Contribution:} 
In this work, we focus on the challenging locomotion problem of high-speed running under significant disturbances in the form of added loads and terrain variability. Addressing both speed and robustness together in the same framework is difficult due to the trade-off that must occur between speed and maximum load-carrying capability. Concurrent work has studied locomotion at various velocities, for example with maximum 1.2 $m/s$ on ANYmal in~\cite{hwangbo2019anymal}, and 3.75 $m/s$ on Mini-Cheetah in~\cite{ji2022concurrent} without disturbances. Robustness to environmental conditions or robot mass was also studied on A1 in\cite{kumar2021rma}, however the maximum velocity was 0.35 $m/s$. In contrast, we make significant advances in combining both of these areas, achieving 2 $m/s$ with a 5 $kg$ load representing 42\% of the nominal robot mass while bounding with a natural gait. To the best of our knowledge, this is the highest velocity and load combination so far demonstrated on A1.

We present a framework for learning fast and robust running controllers directly in Cartesian space, where the policy chooses desired end effector positions for each quadruped leg. Joint torques are then computed with Cartesian space PD control. This action space provides more structure than learning in joint space, including a more obvious action mapping for the agent, yet avoids biases introduced from trajectory generators. Force control and Cartesian space control have previously been successfully used for legged robots in both purely model-based control methods \cite{dicarlo2018mpc,bellicoso2018dynamic,kim2019highly,sombolestan2021adaptive} as well as when combined together with learning approaches for stance-control~\cite{yang2022fast,xie2021glide}. Since such methods show robustness to rough terrains, we believe that adopting this idea to more vanilla learning approaches helps to retain this robustness property, without over-specifying or biasing the solution. Results show benefits from this approach include: 
\begin{itemize}
    \item minimal needed reward shaping
    \item improved sample efficiency even under significant model uncertainty
    \item the emergence of high-speed, robust, and natural gaits such as bounding and galloping
    \item robust bounding over various terrains and adaptability to unknown loads of up to 5kg ($42\%$ of the robot's mass)
\end{itemize}

The rest of this paper is organized as follows. Section~\ref{sec:background} provides background details on reinforcement learning. Section~\ref{sec:method} describes our design choices and training set up for learning fast and robust dynamic locomotion in Cartesian space. Section~\ref{sec:result} shows results from learning our controller, sim-to-sim and sim-to-real transfers, and comparison with other approaches. Lastly, a brief conclusion is given in Section~\ref{sec:conclusion}.

%%%%%%%%%%%%%%%%%%%%%%%%%%%%%%%%%%%%%%%%%%%%%%%%%%%%%%%%%%%%%%%%%%%%%%%%%%%%%%%%
% Background
%%%%%%%%%%%%%%%%%%%%%%%%%%%%%%%%%%%%%%%%%%%%%%%%%%%%%%%%%%%%%%%%%%%%%%%%%%%%%%%%
\section{Background}
\label{sec:background}

\subsection{Reinforcement Learning}
In the reinforcement learning framework~\cite{sutton1998rl}, an agent interacts with an environment modeled as a Markov Decision Process (MDP). An MDP is given by a 4-tuple $(\mathcal{S,A,P,R})$, where $\mathcal{S}$ is the set of states, $\mathcal{A}$ is the set of actions available to the agent, $\mathcal{P}: \mathcal{S} \times \mathcal{A} \times \mathcal{S} \rightarrow \mathbb{R}$ is the transition function, where $\mathcal{P}(s_{t+1} | s_t, a_t)$ gives the probability of being in state $s_t$, taking action $a_t$, and ending up in state $s_{t+1}$, and  $\mathcal{R}: \mathcal{S} \times \mathcal{A} \times \mathcal{S} \rightarrow \mathbb{R}$ is the reward function, where $\mathcal{R}(s_t,a_t,s_{t+1})$ gives the expected reward for being in state $s_t$, taking action $a_t$, and ending up in state $s_{t+1}$. The goal of an agent is to interact with the environment by selecting actions that will maximize future rewards.

There are several popular algorithms for determining the optimal policy $\pi$ to maximize the expected return, and while we expect any RL algorithm to work in our framework, in this paper we use Proximal Policy Optimization (PPO)~\cite{ppo}, a state-of-the-art on-policy algorithm for continuous control problems, which optimizes the following clipped surrogate objective: 
\begin{align}
L^{CLIP}(\theta) = \hat{\mathbb{E}}_t [\min(r_t(\theta)\hat{A}_t, \text{clip}(r_t(\theta),1-\epsilon,1+\epsilon)\hat{A}_t)] \nonumber
\end{align}
where $\hat{A}_t$ is an estimator of the advantage function at time step $t$ as in~\cite{Schulman15}, and $r_t(\theta)$ denotes the probability ratio
\begin{align}
r_t(\theta) = \frac 
			{{\pi}_{\theta}(a_t | s_t)} 
            {{\pi}_{\theta_{old}}(a_t | s_t)} \nonumber
\end{align}
where $\pi_\theta$ is a stochastic policy, and $\theta_{old}$ is the vector of policy parameters before the update.
This objective seeks to penalize too large of a policy update, which means penalizing deviations of $r_t(\theta)$ from 1.

%%%%%%%%%%%%%%%%%%%%%%%%%%%%%%%%%%%%%%%%%%%%%%%%%%%%%%%%%%%%%%%%%%%%%%%%%%%%%%%%
% Method 
%%%%%%%%%%%%%%%%%%%%%%%%%%%%%%%%%%%%%%%%%%%%%%%%%%%%%%%%%%%%%%%%%%%%%%%%%%%%%%%%
\section{Learning Robust Locomotion Controllers}
\label{sec:method}

\begin{figure}[!tpb]
      \centering
      \vspace{0.06in}
    \includegraphics[width=\linewidth]{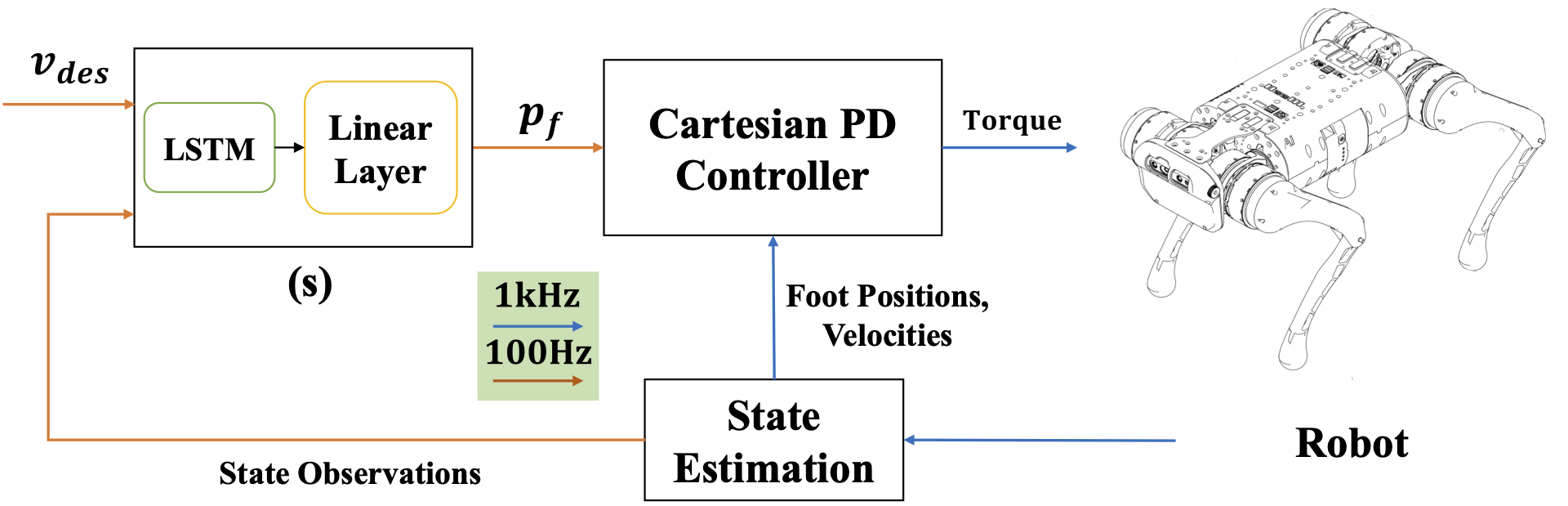}
      \caption{\textbf{Method Overview:} The output of the neural network policy $\bm{\pi(s)}$ are the desired foot position commands $\bm{p}_{f}$ (updated at 100 Hz), which are sent to a Cartesian PD controller (updated at 1 kHz) to control each foot in the leg frame. }
      \label{fig:overview}
      \vspace{-0.6cm}
\end{figure}

In this section we describe our reinforcement learning framework and design decisions for learning fast and robust locomotion controllers for quadruped robots. The control diagram is presented in Figure~\ref{fig:overview}, and we explain all components below. 

%-------------------------------------Action Space-------------------------------------
\subsection{Action Space}
We propose learning desired foot positions in Cartesian space, giving an action space of $\vec{a} \in \mathbb{R}^{12}$, which will then be tracked with Cartesian PD control. Compared with learning in joint space, this gives a naturally limited action space with intuitive ranges that produces an exact mapping to the environment. Additionally, unlike approaches using inverse kinematics, such choice of action space and control allows us to apply implicit force control to track the foot motion instead of joint PD control.

More precisely, the agent chooses Cartesian foot positions in the leg frame from each hip location in the range $[-0.2,0.2]\ m$ for $x$, $[-0.05,0.05]\ m$ for $y$, and $[-0.33,-0.15]\ m$ for $z$, which the policy selects at 100 Hz. These ranges are chosen to allow for speed in the body $x$ direction, small corrections in the body $y$ direction, as well as maintaining a minimum height and clearance of obstacles in the $z$ direction. The foot positions for each leg are then tracked with Cartesian PD control at 1 kHz with:
\begin{eqnarray}
    \bm{\tau} = \bm{J}(\bm{q})^\top \left[ \bm{K}_p \left(\bm{p}_d - \bm{p} \right) - \bm{K}_d \left( \bm{v} \right)  \right] 
\end{eqnarray}
where $\bm{J}(\bm{q})$ is the foot Jacobian at joint configuration $\bm{q}$,  $\bm{p}_d$ are the desired foot positions learned with RL, $\bm{p}$ and $\bm{v}$ are the current foot positions and velocities in the leg frame, and $\bm{K}_p$ and $\bm{K}_d$ are diagonal matrices of proportional and derivative gains in Cartesian coordinates. We note it is possible to successfully learn locomotion controllers for a wide range of Cartesian PD gains, but we present results for $(\bm{K}_{p}=500\bm{I}_3,\ \bm{K}_{d}=10\bm{I}_3)$. To achieve smooth and stable motions, we also incorporate a small joint damping gain $\bm{K}_{p,joint} = 5$ for each motor. 

As is common in reinforcement learning implementations, our policy network outputs actions in the range $[-1,1]$, which are then scaled to the ranges discussed above.

%-------------------------------------Observation Space-------------------------------------
\subsection{Observation Space}
The observation space consists of the velocity command along with parameters that can be directly measured or estimated from our quadruped's onboard sensors: body state (position, orientation, linear and angular velocities), joint state (positions, velocities), foot state (positions, velocities).

%------------------------------------- Reward-------------------------------------
\subsection{Reward Function} 

We design a reward function with the goal of learning fast locomotion policies while encouraging energy efficiency and not falling down. In our simulation studies, we find that a simple reward function consisting only of velocity tracking $v_{base}$ and a penalty on energy is sufficient for learning fast and natural gaits. However, for improved sim-to-real transfer, we add additional terms as shown in Table \ref{tab:reward}. This reward function focuses on task space costs such as the robot's linear velocity, orientation, lateral drift, and height, while also penalizing energy consumption. Although we do not specify a gait clock in our framework, we add a term to encourage longer swing phases, which along with the terrain randomization (Section~\ref{sec:terrain}) results in the agent learning better foot clearance to traverse uneven terrain. 

At each episode reset, we randomly sample a new desired $v_{base}$ command in the range of 0.7 to 4 $m/s$ to learn a continuous policy. 
An episode terminates after 1000 time steps, corresponding to 10 seconds, or if the quadruped loses balance and falls, in which case the agent receives a penalty of $-10$. 

\begin{table}[tpb]
\centering
\vspace{0.06in}
\caption{Reward function terms. $v$ is the linear velocity; $t_i^k$ is the time (in seconds) when foot $i$ touches the ground for the $k^{th}$ time; $\bm{\omega}$ is the base orientation represented as a quaternion; $y$ and $z$ are base coordinates.}
\begin{tabular}{ c c c }
Name & Formula & Weight \\
\hline
Velocity reward & 1 - $|v_{base} - v_{des}|$ & 0.1 \\
Feet swing reward & $\sum_{i=0}^3 (t_i^k - t_i^{k-1} - 0.5)$ & 0.2 \\
Energy penalty & $\int_{t}^{t+1} |\bm{\tau} \cdot \dot{\bm{q}} | dt$ & -0.008 \\
Orientation penalty & $\| \bm{\omega} - (0,0,0,1) \|$ & -0.1 \\
Lateral drift penalty & $|y|$ & -0.1 \\
Height penalty & $|z - 0.3|$ & -0.1 \\
\hline
\end{tabular} \\
\label{tab:reward}
\vspace{-1.5em}
\end{table}

%------------------------------------- Environment Details -------------------------------------
\subsection{Dynamics Randomization}
\label{sec:env_details}

Similarly to the previously discussed studies in Section~\ref{sec:introduction} on learning robust controllers with deep reinforcement learning, we make use of dynamics randomization toward avoiding the exploitation of simulator dynamics, and to ease the sim-to-real transfer. Our dynamics randomization choices are summarized in Table~\ref{table:dyn_rand}, and we discuss them below. 

At the start of each environment reset, the mass of each body part (base, hips, thighs, calves, feet) is individually perturbed randomly by up to 20\% of its nominal value, and the inertia is set accordingly. The coefficient of friction of the ground (and any random objects added to the simulation) is also randomly chosen in [0.5, 1]. 

For added robustness, and to test the capabilities of the system, a random mass is attached to the base with probability 0.8. This mass is uniformly chosen to be in [0, 5] $kg$, and randomly placed within [0.15, 0.05, 0.05] $m$ offsets from the base center of mass. We also introduce Gaussian noise to the observation space with a standard deviation of 0.05. 

\subsection{Terrain Randomization}
\label{sec:terrain}
Since our reward function doesn't consider swing foot clearance, our framework generates random terrain to train the policy so that the resulting controller can command a proper clearance for the swing foot and adapt to rough terrain.  At the start of each environment reset, we generate 100 random boxes up to 0.02 $m$ high, up to 1 $m$ in length and width, and with random orientation. These are randomly distributed in a 20 $\times$ 6 $m$ grid in front of the agent, and walls are added at $\pm 3\ m$ on the $y$ axis so the agent cannot learn to avoid the random boxes.  With this setup, the robot learns to lift its feet up high enough to overcome rough terrain.

%------------------------------------- Model Details -------------------------------------
\subsection{Network Representation}
\label{sec:model}
We tested neural network architectures consisting of (a) purely multi-layer perceptrons (MLPs) as well as (b) a Long Short-Term Memory (LSTM) layer followed by a fully connected layer. While both architecture types were successful in producing locomotion gaits with our framework in simulation, we found that the architecture including an LSTM layer transferred more successfully to the hardware. 

In our final architecture as shown in Figure~\ref{fig:overview}, the LSTM layer contains one hidden layer with a hidden unit size of 128, and the fully connected layer has a dimension of 128 neurons. The LSTM layer first takes observation data collected from robot sensors and state estimation to extract latent features. The fully connected layer then maps the LSTM output to final actions. In addition, the LSTM output is shared to a second fully connected layer to represent the critic network.

\begin{table}[tpb]
\centering
\vspace{0.06in}
\caption{Randomized physical parameters and their ranges.}
\begin{tabular}{ c c c }
Parameter & Lower Bound & Upper Bound \\
\hline
Mass (each body link) & 80\% & 120\% \\
Added mass & 0 $kg$ & 5 $kg$ \\
Added mass base offset & [-0.15,-0.05,-0.05] $m$ & [0.15,0.05,0.05] $m$ \\
Coefficient of friction & 0.5 & 1 \\
\hline
\end{tabular} \\
\label{table:dyn_rand}
\vspace{-1.0em}
\end{table}

\begin{table}[tpb]
\centering
\vspace{0.06in}
\caption{PPO Hyperparameters.}
\begin{tabular}{ c c  }
Parameter & Value \\
\hline
Horizon (T) & 4096 \\
Optimizer & Adam  \\
Learning rate & $1 \cdot 10^{-4}$ \\
Number of epochs & 10 \\
Minibatch size & 128 \\
Discount ($\gamma$) & 0.99 \\
GAE prarmeter ($\lambda$) & 0.95 \\
Clipping parameter ($\epsilon$) & 0.2 \\
VF coeff. ($c_1$) & 1 \\
\hline
\end{tabular} \\
\label{table:tab1}
\vspace{-1.5em}
\end{table}

%%%%%%%%%%%%%%%%%%%%%%%%%%%%%%%%%%%%%%%%%%%%%%%%%%%%%%%%%%%%%%%%%%%%%%%%%%%%
% Results
%%%%%%%%%%%%%%%%%%%%%%%%%%%%%%%%%%%%%%%%%%%%%%%%%%%%%%%%%%%%%%%%%%%%%%%%%%%%
\section{Results}
\label{sec:result}

\begin{figure}[!tpb]
      \vspace{0.06in}
      \centering
      \includegraphics[width=1.7in,trim={0 0 0 2cm},clip]{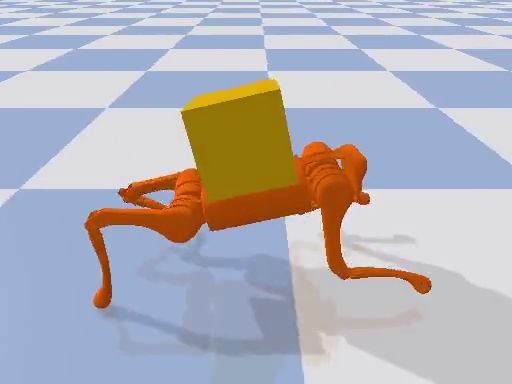}\includegraphics[width=1.7in,trim={0 0 0 2cm},clip]{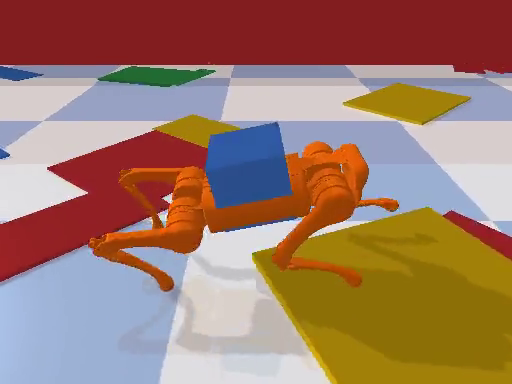}
      \includegraphics[width=3.4in]{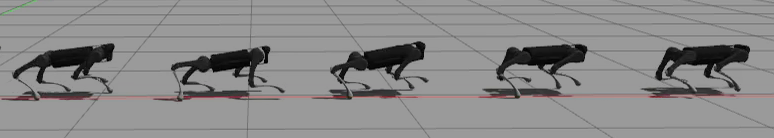} \\
      \includegraphics[width=3.4in]{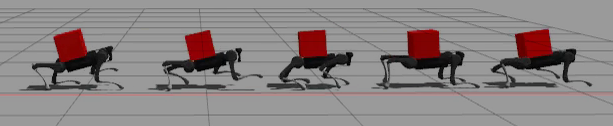} \\
      \caption{Dynamic running performance in PyBullet~\cite{pybullet}, and sim-to-sim transfer to Gazebo. \textbf{Top:} Running with a 10 $kg$ load (83\% of nominal mass) on flat terrain (left), and a 6 $kg$ load (50\% of nominal mass) over rough random terrain up to 0.04 $m$ in height (right). \textbf{Middle:} 1 second of snapshots in Gazebo, running at 4 $m/s$.
      \textbf{Bottom:} 1 second of snapshots in Gazebo, running at 3.5 $m/s$ with a 10 $kg$ mass. 
      Simulation video results can be found at \insertYoutubeLink. 
      }
       \vspace{-1.5em}
      \label{fig:results_sim}
\end{figure}

\begin{figure*}[t]
    \vspace{0.06in}
    \centering
    \includegraphics[width=\linewidth]{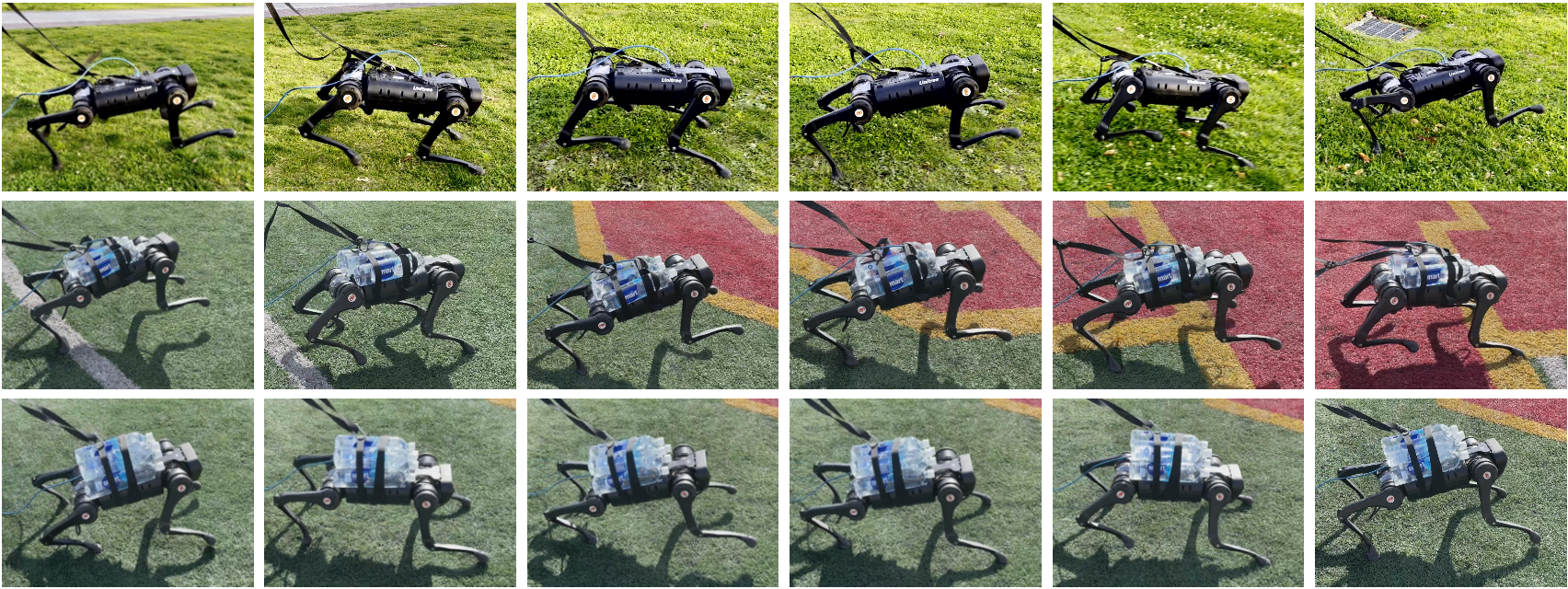}
    \caption{Motion snapshots of A1 running on rough terrain (top), with a 3 kg load (middle), and with a 5 kg load (bottom). }
    \label{fig:snapshots}
   \vspace{-1.5em}
\end{figure*}

In this section we discuss results from using our method to learn fast and robust locomotion controllers. We use PyBullet~\cite{pybullet} as the physics engine for training and simulation purposes, and the Unitree A1 quadruped model~\cite{unitreeA1}. All policies are learned with PPO~\cite{ppo}, and our neural networks are explained in Section~\ref{sec:model}. Other training hyperparameters are listed in Table~\ref{table:tab1}. 

Figure~\ref{fig:results_sim} shows snapshots from executing the learned policy in the training simulator with a 10 $kg$ load (top left), and over rough terrain with blocks of up to 0.04 $m$ high and a load of 6 $kg$ (top right). 
The bottom of Fig.~\ref{fig:results_sim} shows the results of a sim-to-sim transfer to Gazebo using the ODE physics engine, with and without running with a 10 $kg$ mass. 

We also present snapshots from sim-to-real transfers in Figures~\ref{fig:intro} and~\ref{fig:snapshots}. The reader is encouraged to watch the supplementary video for clearer visualizations. 

For our experiments, we are specifically interested in the following questions: 
\begin{itemize}
    \item How does environmental noise affect the agent's ability to learn, as well as resulting policies and gaits?
    \item Which environmental noise parameters are most critical for a successful sim-to-sim or sim-to-real transfer? 
    \item How does action space choice affect the sim-to-sim and sim-to-real transfer? 
\end{itemize}

%-------------------------------------------------------------------------
\subsubsection{Training curves}
\begin{figure}[!tpb]
    \centering
    \includegraphics[width=3.4in]{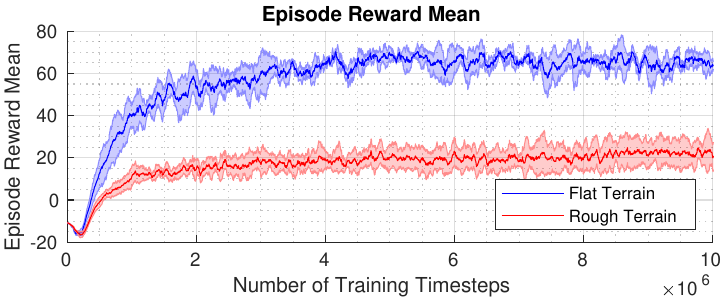} \\
    \caption{Episode reward mean while training in our proposed action space. We compare training over flat terrain only and randomly varying rough terrain, which consists of random boxes up to 0.02 $m$ in height.}
    \label{fig:training_curves}
     \vspace{-1.5em}
\end{figure}

Figure~\ref{fig:training_curves} shows training curves for the environment described in Section~\ref{sec:env_details} over flat terrain, as well as for randomly varying rough terrain as described in Section~\ref{sec:terrain}. In both cases, results show that training converges within a few million timesteps on both flat and rough terrain. 

There is a classic trade-off between speed and robustness on the rough terrain, as the agent needs to lift its feet higher to clear the terrain, which necessitates expending more energy as well as added difficulty of tracking the commanded forward velocity. Also, we note that the agent is still likely to catch its feet on the sides of the boxes, but is able to recover in most cases if the added mass is not too large, or if the speed is not too high. As a reactive controller can only do so well in such instances, in future work we plan to incorporate exteroceptive measurements such as vision to run over even rougher terrain.  

%-------------------------------------------------------------------------
\subsubsection{Effects of Dynamics Randomization} 
To evaluate the requirement and usefulness of our dynamics randomization choices, we compare policies trained with systematically increasing environment noise parameters in PyBullet, and their performance at test time in a sim-to-sim transfer. Each trained policy is run in Gazebo for 10 trials, where a trial lasts for a maximum of 20 seconds and terminates early due to a fall, and we evaluate the mean distance the robot is able to run in this time. For these tests we use the nominal quadruped model (without any noise in the mass/inertia, nor added load) in Gazebo, and we load and query our trained policy network with the TensorFlow C++ API.  

Results from various training noise are summarized in Table~\ref{table:sim2sim_dynrand}. Notably, with full randomization as described in Section~\ref{sec:env_details}, we achieve a 100\% success rate without any falls, where the agent consistently runs just under 80 $m$ in 20 seconds. With using any subset of the dynamics randomization, we observe that the quadruped always falls before the 20 seconds elapse. In these cases, the distance traveled is estimated to be just prior to the base hitting the ground. 
As the significantly varying added load adds the most randomization, it is intuitive that this training noise is most helpful for the sim-to-sim (and sim-to-real) transfer.  
%%%-------------------------------------------------------------------
\begin{table}[tpb]
\centering
\caption{Sim-to-sim success rate for 10 trials from training with varying dynamics randomization. A successful trial represents running without falling for 20 seconds.  }
\begin{tabular}{ c c c }
Dynamics Randomization & Mean Dist. (m) & Success Rate \\
\hline
None & $<2$ & 0\\
Varying $\mu$ only & 5 & 0 \\ 
Varying $\mu$, added load & 7 & 0 \\
Varying $\mu$, 10\% mass & 5 & 0 \\
Varying $\mu$, 20\% mass & 4 & 0 \\
Varying $\mu$, 10\% mass, added load & 14 & 0 \\
Varying $\mu$, 20\% mass, added load & \textbf{78.9} & \textbf{1}\\ 
\hline
\end{tabular} \\
\label{table:sim2sim_dynrand}
\vspace{-1.5em}
\end{table}

%-----------------------------------------------------------------
\subsubsection{Comparison with Joint PD} 
We compare training in Cartesian space with Cartesian PD control with training directly in joint space. Recent works have found that with proper reward shaping and dynamics randomization, policies trained in joint space can transfer well from simulation to hardware~\cite{hwangbo2019anymal,lee2020anymal,tan2018minitaur}. However, such works have focused on robustness, and speeds lower than those attainable with our quadruped. 

We tested training fast running policies in joint space with a range of joint gains, namely $k_p \in [20,100]$ and $k_d \in [0.1, 1]$, using the same observation space, dynamics randomization, and simple reward function from Table~\ref{tab:reward}. We trained at least 3 policies for each set of gains tested, and also tested varying the action space to limit the joint ranges to those seen when training in Cartesian space, toward ensuring a fair comparison. 
For all combinations tested, the agent is able to learn to run without falling in PyBullet, for the same speeds as with training in Cartesian space. However, almost all policies converge to unnatural-looking gaits, where one or both rear legs extend outward and back, which appear to be used more for balancing. This behavior remains the same even if we limit the action space to the joint ranges used by policies trained in Cartesian space. 

Although still able to successfully run in PyBullet, the policies learned in joint space do not transfer well to another simulator. The unnatural looking gaits may be the result of the agent getting stuck in local minima which still achieve the same reward through exploiting the simulator dynamics, due to the simple reward function that does not over-specify ``good'' behavior. We summarize the best results obtained for policies trained with various joint gains in Table~\ref{table:jointpd}. We run 10 trials for each policy in the transfer to Gazebo, and record the average distance traveled in 20 seconds, as well as the success rate of not falling. 

\begin{table}[tpb]
\centering
\vspace{0.06in} 
\caption{Sim-to-sim performance of 10 trials from policies learned in joint space with various joint PD gains. All joint gains can produce fast and stable (though often unnatural-looking) locomotion in PyBullet, but do not transfer well to another simulator. }
\begin{tabular}{ |c | c | c | c |  }
\hline 
$k_p$ & $k_d$ &  Mean Dist.  &  Success Rate\\[.2ex]
\hline
30 & 0.1 & 59 & 0.9 \\
50 & 0.1 & 20 & 0.2 \\
20  & 0.5 & 11 & 0 \\
30 & 0.5 & 24 & 0.1 \\
40 & 0.5 & 56 & 0.7 \\
50 & 0.5 & 37 & 0.9 \\
60 & 0.5 & 26  & 0.7 \\
70 & 0.5 & 15 & 0.4 \\
50  & 1 & 36 & 0.4 \\
60 & 1 & 24  & 0.3 \\
\hline
\end{tabular} \\
\label{table:jointpd}
 \vspace{-0.5em}
\end{table}

%%%-------------------------------------------------------------------

\begin{figure}[tpb]
    \centering
    \includegraphics[width=3.4in]{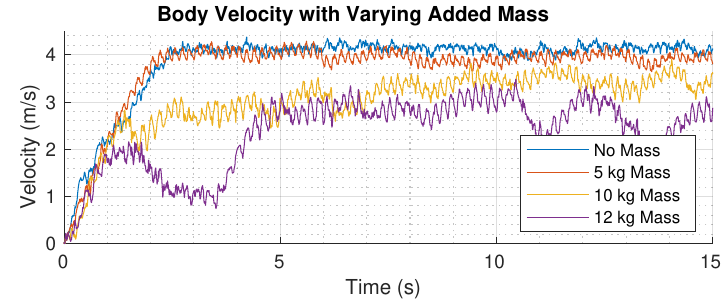} \\
    \caption{Body velocity in Gazebo with varying added masses from sample policy executions.}
    \label{fig:vels_mass}
    \vspace{-0.5em}
\end{figure}

%------------------------------------------------------------------
\subsubsection{Performance with Varying Loads} 

We next test the full performance in the sim-to-sim transfer of the policy learned in Cartesian space as described in Section~\ref{sec:method}. 
Figure~\ref{fig:vels_mass} shows the body velocity from sample policy executions in Gazebo with varying added masses. With 5 $kg$ of added mass (42\% of the nominal robot mass), there is little noticeable difference in performance from the ideal case. As the added load mass increases even further, it becomes more difficult for the agent to maintain higher speeds, though still possible to make forward progress. 

\begin{figure}[!tpb]
    \centering
    \includegraphics[width=3.4in]{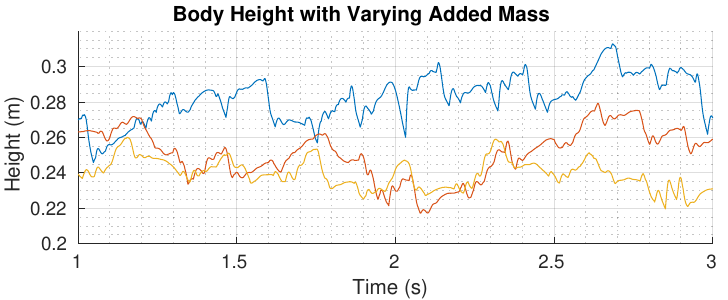}\\
    \includegraphics[width=3.4in]{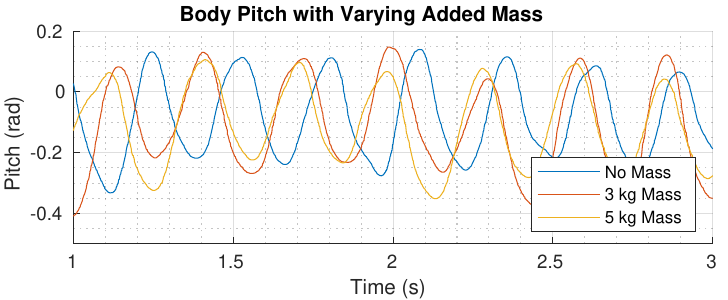}
    \caption{Body height and orientation for running hardware experiments with different loads.}
    \label{fig:height_rp}
    \vspace{-1.7em}
\end{figure}

\subsubsection{Sim-to-Real Transfer} 
We successfully transfer the learned LSTM policies onto the Unitree A1 robot hardware. Figure~\ref{fig:snapshots} shows snapshots from the robot robustly running over rough terrain with a natural-looking bounding gait, including adaptability to added loads up to 5 kg. 

Our video shows additional results of running on various terrains (rough and variable grass including rocks, pavements, uneven terrain) including under heavy loads (42\% of the nominal robot mass). We would like to emphasize that the bumpy grass field shown in Figure~\ref{fig:snapshots} is actually very uneven. In addition, since the policy is blind, it is challenging to account for high obstacles at high speeds. Therefore, as part of our future work, we are working on adding vision input into the learning framework to allow high speed running on more challenging terrains. 

In Figure~\ref{fig:height_rp}, we compare the body height and orientation while running with no load, with a 3 $kg$ load, and with a 5 $kg$ load. The plots show 3 seconds of time when the agent has reached a steady-state velocity in all cases. The height of the quadruped is lower when carrying heavier loads, though the pitch remains similar. 

As we are able to achieve higher velocities in sim-to-sim (4 $m/s$) than sim-to-real (2 $m/s$), we believe there are several factors for this discrepancy. Unlike works focusing on lower speeds, our work on high-speed running is expected to encounter a higher level of model uncertainty between simulation and the real robot hardware. In simulation, the motors are not being modeled (i.e. no delay or battery depletion), and therefore it is often the case that the commanded torque cannot be achieved on the hardware (as it can be in simulation). Simulation also does not capture motor dynamics at the boundary of maximum motor torque and speed. The contact dynamics will also be much more inaccurate under highly dynamic motions with large and variable ground impacts (for example the unmodeled deformable feet of A1 undergo significant deformation during stance phase, especially at high speeds). Finally, the state estimation on the hardware is very noisy compared to in simulation.

%%%%%%%%%%%%%%%%%%%%%%%%%%%%%%%%%%%%%%%%%%%%%%%%%%%%%%%%%%%%%%%%%%%%%%%%%%%%%%%%
% Conclusion
%%%%%%%%%%%%%%%%%%%%%%%%%%%%%%%%%%%%%%%%%%%%%%%%%%%%%%%%%%%%%%%%%%%%%%%%%%%%%%%%
\section{Conclusion}
\label{sec:conclusion}

In this work, we presented a framework for learning robust quadruped running, with the emergence of natural bounding gaits through the agent selecting actions in Cartesian space and tracking them with Cartesian PD control. 
Compared to other recent works, our results showed minimal needed reward shaping, improved sample efficiency, the emergence of natural gaits such as galloping and bounding, and a straightforward sim-to-real transfer. We also showed robustness to environmental disturbances such as unperceived, uneven terrain as well as unknown added masses representing 42\% of the nominal mass of the robot on hardware. In addition to previously proposed dynamics randomization, we find that training with unknown added loads can play a significant role in successful sim-to-sim and sim-to-real transfers.  
In future work, we plan to incorporate exteroceptive measurements such as vision to improve dynamic capabilities over rougher terrain. 

%%%%%%%%%%%%%%%%%%%%%%%%%%%%%%%%%%%%%%%%%%%%%%%%%%%%%%%%%%%%%%%%%%%%%%%%%%%%%%%%
\bibliographystyle{IEEEtran}
\bibliography{refs} 
\end{document}